\documentclass[journal]{IEEEtran}

\ifCLASSINFOpdf
\else
\fi

\usepackage[utf8]{inputenc} 
\usepackage[T1]{fontenc}    
\usepackage{hyperref}       
\usepackage{url}            
\usepackage{booktabs}       
\usepackage{amsfonts}       
\usepackage{nicefrac}       
\usepackage{microtype}      
\usepackage{graphicx}
\usepackage{amsmath}
\usepackage{adjustbox}
\usepackage{bm,algorithm,algorithmic}
\DeclareMathOperator{\KL}{KL}

\begin{document}
%
\title{Uncertainty Quantification in  Deep MRI Reconstruction}
%
%
%

\author{Vineet Edupuganti,
        Morteza Mardani, Shreyas Vasanawala, and John Pauly
\thanks{V. Edupuganti, M. Mardani, and J. Pauly are with the Department
of Electrical Engineering, Stanford University, Stanford,
CA, 94305 USA. e-mails: \textit{ve5, morteza, pauly@stanford.edu.} }
\thanks{S. Vasanawala is with the Department of Radiology, Stanford University, Stanford,
CA, 94305 USA. e-mail: \textit{vasanawala@stanford.edu.} }
}

\bstctlcite{IEEEexample:BSTcontrol}

\markboth{Submitted to IEEE TRANSACTIONS ON MEDICAL IMAGING}{ }

\maketitle

\begin{abstract}
Reliable MRI is crucial for accurate interpretation in therapeutic and diagnostic tasks. However, undersampling during MRI acquisition as well as the overparameterized and non-transparent nature of deep learning (DL) leaves substantial uncertainty about the accuracy of DL reconstruction. With this in mind, this study aims to quantify the uncertainty in image recovery with DL models. To this end, we first leverage variational autoencoders (VAEs) to develop a probabilistic reconstruction scheme that maps out (low-quality) short scans with aliasing artifacts to the diagnostic-quality ones. The VAE encodes the acquisition uncertainty in a latent code and naturally offers a posterior of the image from which one can generate pixel variance maps using Monte-Carlo sampling. Accurately predicting risk requires knowledge of the bias as well, for which we leverage Stein's Unbiased Risk Estimator (SURE) as a proxy for mean-squared-error (MSE). Extensive empirical experiments are performed for Knee MRI reconstruction under different training losses (adversarial and pixel-wise) and unrolled recurrent network architectures. Our key observations indicate that: 1) adversarial losses introduce more uncertainty; and 2) recurrent unrolled nets reduce the prediction uncertainty and risk.





\end{abstract}


%
\IEEEpeerreviewmaketitle

\section{Introduction}
\begin{figure}
\begin{center}
	\includegraphics[width = 8.2cm] {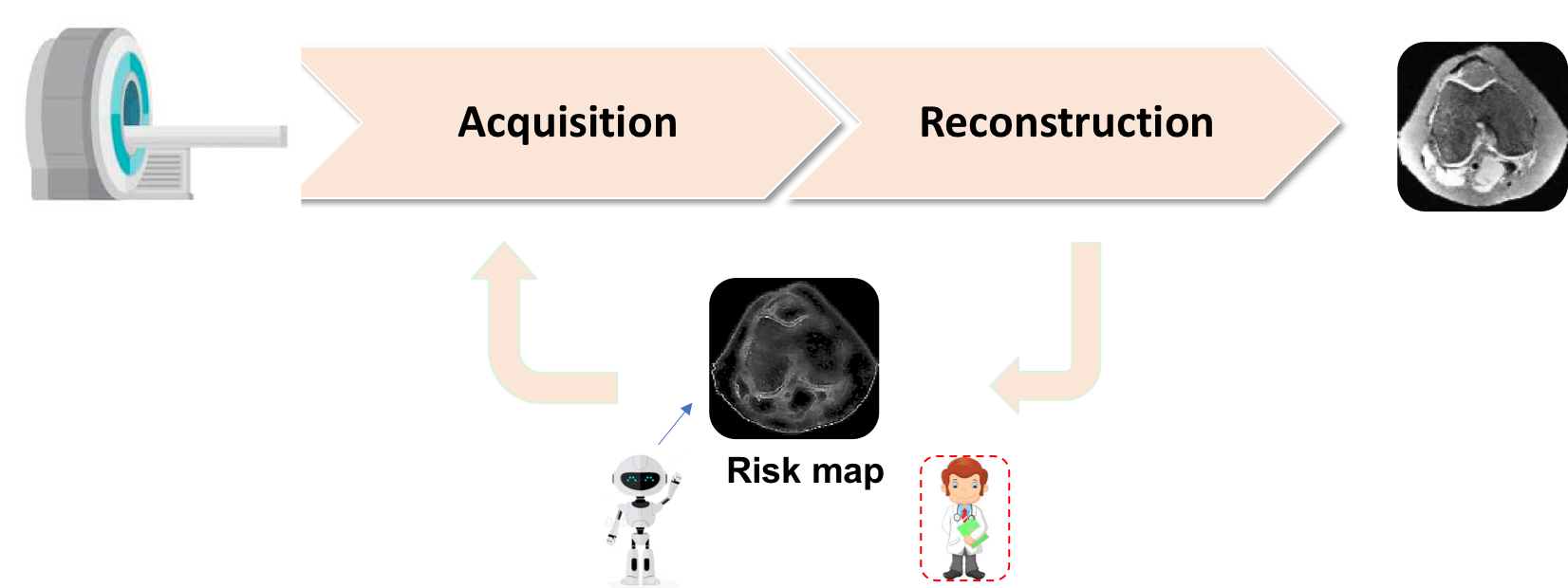}
    \caption{\small{Processing pipeline for clinical operation. The scanner acquires k-space data, followed by image 
reconstruction that is subject to artifacts. Traditionally, physicians are in the loop to assess the image quality which is labor intensive and error prone. }}
    \label{fig:workflow}
\end{center}
\end{figure}

\IEEEPARstart{A}{rtificial} intelligence (AI) has introduced a paradigm shift in medical image reconstruction in the last few years, offering significant improvements in speed and image quality \cite{radford2015unsupervised, quan2017compressed, zhu2018image, lee2017deep, bora2017compressed, mousavi2015deep}. However, the prevailing methods for image reconstruction leverage historical patient data to train deep neural networks, and then use these models on new unseen patient data. This practice raises concerns about \textit{generalization}, given that predictions can be biased towards the previously-seen training data, and consequently may struggle to detect novel pathological details in the test subjects. This is deeply problematic for both patients and physicians and can potentially alter diagnoses. It is thus crucial to ensure that reconstructions are accurate for new and unseen subjects. This well motivates developing effective and automated risk assessment tools for monitoring image reconstruction algorithms as a part of the imaging pipeline (see the clinical workflow in Fig. \ref{fig:workflow}).






Despite the importance of assessing risk in medical image reconstruction, little work has explored the robustness of deep learning (DL) architectures in inverse problems. Not only is the introduction of image artifacts fairly common using such models, but there do not currently exist suitable empirical methods that enable the quantification of uncertainty \cite{mardani2018deep}. Given the pernicious effects that the presence of image artifacts can have, such methods could have utility both as an evaluation metric and as a way of gaining interpretability regarding risk factors for a given model and dataset \cite{kendall2017uncertainties}.


To this end, this work introduces procedures that can provide insights into the robustness of DL MR reconstruction schemes. In doing so, we develop a variational autoencoder (VAE) model for MR image recovery, which is notable for its low error and probabilistic nature which is well-suited to an exploration of model uncertainty. We first generate admissible reconstructions under an array of different model conditions. We then utilize Monte Carlo sampling methods to better understand the variations and errors in the output image distribution under various model settings \cite{harrell1996multivariable}. Finally, because accurately assessing bias and error with the Monte Carlo approach requires access to the true (fully-sampled) image, we leverage the Stein's Unbiased Risk Estimator (SURE) analysis of the DL model, which serves as a surrogate for mean squared error (MSE) even when the ground truth is unknown. 

Extensive empirical evaluations are performed on real-world medical MR image data. Our key observations include: 1) using an adversarial loss function introduces more pixel uncertainty than a standard pixel-wise loss, while better recovering the high-frequencies; 2) a cascaded network architecture better leverages the physical constraint of the problem and results in higher confidence reconstructions 3) SURE effectively approximates MSE and serves as a valuable tool for assessing risk when the ground truth reconstruction (i.e. fully sampled image) is unavailable.

The major contributions of this paper are:

\begin{itemize}
    \item A novel VAE scheme for learning inverse maps that produces pixel uncertainty maps
    \item Quantifying risk using SURE by taking into account the end-to-end network Jacobian 
    \item Extensive evaluations and statistical analysis of uncertainty for various network architectures and training losses

\end{itemize}

The rest of this paper is organized as follows: Section II reviews relevant literature. Section III introduces the preliminaries on neural recovery algorithms and states the problem. Section IV details the VAE architecture. Section V explains the Monte Carlo and SURE approaches to quantifying uncertainty, including a density compensation step that makes the use of SURE more practical. Empirical evaluations are then reported in Section VI, and Section VII discusses the conclusions and future directions.

\section{Related Work}
Medical image reconstruction methods rooted in deep learning have been widely explored. Of the various model architectures that have been used, adversarial approaches based on generative adversarial networks (GANs) are notable for high reconstruction image quality and the ability to realistically model image details. The generator function for these types of architectures is usually a U-net or ResNet \cite{mardani2018deep, emami2018generating}.

Variational autoencoders (VAEs) have achieved high performance in generating natural images, while also providing probabilistic interpretability of the generation process \cite{larsen2015autoencoding, cai2019multi}. However, this approach has not yet been demonstrated in the realm of medical images.

Meanwhile, a small but growing body of work has examined uncertainty in general computer vision problems. Specifically, measurements of uncertainty have been computed by finding the mean and point-wise standard deviation of test images using Monte Carlo sampling with probabilistic models \cite{mosegaard1995monte}. With such a method, comparing the mean intensities of regions containing an artifact and the surrounding area over several cases can provide statistical insights into the types of errors made by the model \cite{adler2017learning}. 

Other studies have explored using invertible neural networks to learn the complete posterior of system parameters \cite{ardizzone2018analyzing, gilbert2017towards}. Through bootstrapping, point estimate uncertainty can be obtained statistically and analyzed in a manner similar to posterior sampling. Uncertainty has also been analyzed from the standpoint of data rather than variance introduced by generative models in the context of medical imaging \cite{tygert2018compressed}. Techniques such as the bootstrap and jackknife can be used on the sampled input data to produce accurate error maps that provide insight into the most risky ROIs in terms of reconstructions without having access to the ground truth.

Stein's Unbiased Risk Estimator (SURE), which this paper explores as a measure of uncertainty, has also seen some use in imaging applications. Specifically, SURE has been utilized in regularization and for image denoising, where it is explicitly minimized during the optimization step \cite{luisier2007new, deledalle2012unbiased}. However, it has not been widely used in uncertainty estimation or in the context of medical imaging.

Given that most existing approaches of quantifying uncertainty do not apply broadly or are constrained by the chosen models, developing straightforward and effective techniques is very important. Such methods have the potential to enable holistic comparison and evaluation of model architectures across a range of problems, leading to increased robustness in sensitive domains.

\section{Preliminaries and Problem Statement}
\label{submission}
One key application of inverse problems is to Magnetic Resonance Imaging (MRI), where significant undersampling is typically employed to accelerate acquisition, leading to challenging image recovery problems. As a result, the development of accurate and rapid MRI reconstruction methods could enable powerful applications like diagnostic-guided surgery or cost-effective pediatric scanning without anesthesia \cite{hynynen1993mri, vasanawala2011advances}.

In MR imaging, the goal is to recover the true image $x_0 \in \mathbb{C}^n$ from undersampled k-space measurements $y \in \mathbb{C}^m$ with $m \ll n$ that admit 
\begin{align}
    y = \Phi x_0 + v
\end{align}
Here, $\Phi$ includes the sampling mask $\Omega$ and the Fourier operator $F$, as well as coil sensitivities. The noise term $v$ also accounts for measurement noise and unmodeled dynamics. 

\begin{figure} 
\begin{center}
\includegraphics[scale = 0.4] {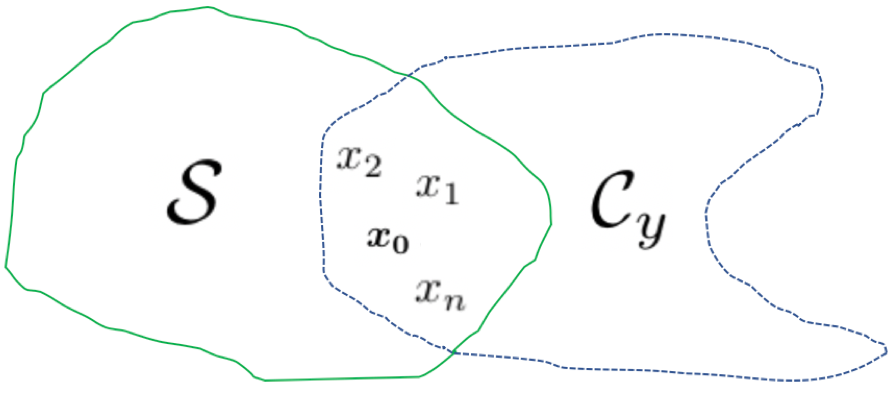}
\caption{\small{Admissible solutions ($x_0$ is the true image).}}
\label{fig:venn_diagram}
\end{center}

\end{figure}

Given the ill-posed nature of this problem, it is necessary to incorporate prior information to obtain high-quality reconstructions. This prior spans across a manifold of realistic images ($\mathcal{S} \subset \mathbb{C}^n)$. However, since not all points on this manifold are consistent with the measurements, we must consider the intersection of the prior manifold $\mathcal{S}$ with a data consistent subspace $\mathcal{C}_y := \{x \in \mathbb{C}^n: y = \Phi x+v\}$ (as shown in Fig. \ref{fig:venn_diagram}). Note that there might be multiple admissible solutions $x_0,x_1, \ldots, x_n$ at the intersection with different likelihoods. 

While DL models can be effectively used for learning the projection onto the intersection $\mathcal{S} \cap \mathcal{C}_y$, performance can be limited when seeing new data unlike the training examples. In particular, one risk is the introduction of realistic artifacts, or so-termed "hallucinations," which can prove costly in a domain as sensitive as medical imaging by misleading radiologists and resulting in incorrect diagnoses \cite{xu2014deep, yang2018dagan}. Hence, analyzing the uncertainty and robustness of DL techniques in MR imaging is essential.


Thus, the objective of this work is to learn a projection onto the intersection between the real image manifold $\mathcal{S}$ and the data consistent subspace $\mathcal{C}_y$ using a DL model, and then evaluate the uncertainty of the model in producing these reconstructions. 


\begin{figure*}
\begin{center}
	\includegraphics[width = 15cm] {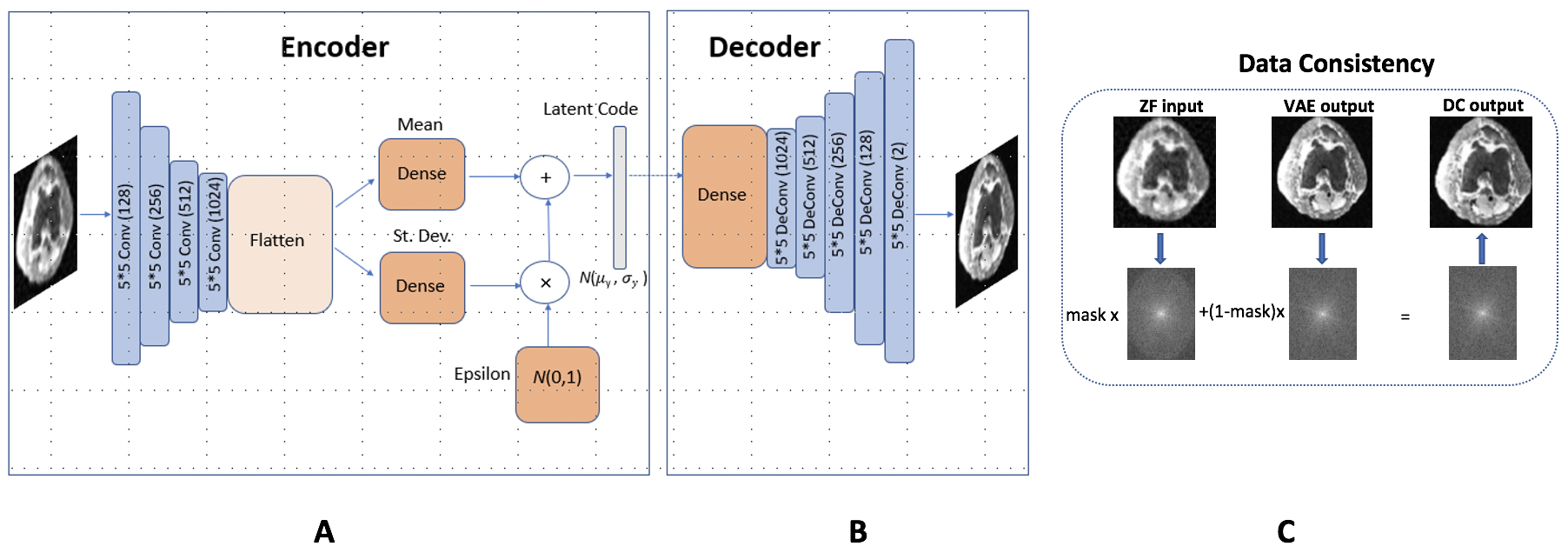}
    \caption{\small{The model architecture, with aliased input images feeding into the VAE encoder (A), the latent code feeding into the VAE decoder (B), and data consistency (C) applied to obtain the output reconstruction. This output can serve as an input to the GAN discriminator (not pictured), which in turn sends feedback to the generator (VAE) when adversarial loss is used. The VAE and data consistency layers are repeated in the case of multiple recurrent blocks. }}
    \label{fig:architecture}
\end{center}
\end{figure*}

\section{VAEs for Medical Image Recovery}
%


For image recovery, we consider a VAE architecture ($h)$, consisting of an encoder $f$ and a decoder $g$ with latent space $z$ as in Fig. \ref{fig:architecture}. While VAEs have been used successfully in low-level computer vision tasks like super-resolution, they have not been applied to medical image recovery. Importantly, the VAE is capable of learning a probability distribution of realistic images that facilitates the exploration process of the manifold $\mathcal{S}$. By randomly sampling latent code vectors corresponding to specific images and enforcing data consistency, we can traverse the space comprising $\mathcal{S}\cap\mathcal{C}_y$ and evaluate the results visually and statistically.


The VAE functions in the following manner. First, the encoder, which takes in input $x_{zf}$, returns $q(z|x)$ (an estimate of the posterior image distribution), where $z$ is an internal latent variable with low dimension. This latent variable formulation is valuable because it enables an expressive family of distributions and is also tractable from the perspective of computing maximum likelihood. Reconstruction occurs through sampling $z$ from $q(z|x)$ and passing the result to the decoder. To facilitate sampling, the posterior is represented by a Gaussian distribution and constrained by a prior distribution $p(z)$, which for simplicity of computation is chosen to be the unit normal distribution.

In generating reconstructions, the VAE balances error with the ability to effectively draw latent code samples from the posterior. As such, the VAE loss function used in training is comprised of a mixture of pixel-wise $\ell_2$ loss and a KL-divergence term, which measures the similarity of two distributions using the expression below

\begin{align*}
\KL(p || q) = \int_{-\infty}^{\infty} p(x) \log \left(\frac{p(x)}{q(x)}\right) dx
\end{align*}

In the VAE loss function, the KL term (which has weight $\eta$) is designed to force the posterior (based on $\mu_x, \sigma_x$ for a given batch) to follow the unit normal distribution of our prior $p(z)$ \cite{dosovitskiy2016generating, kingma2013auto}. As $\eta$ increases, the integrity of the latent code (and thereby ease of sampling after training) is preserved at the expense of reconstruction quality. With reconstruction $\hat{x} = h(x_{zf})$, the training cost used in updating the weights of the model $h$ is formed as

\vspace{-2 mm}
\begin{align*}
\min_{h}     \mathbb{E}_{x,y} \left[ \|\hat{x} - x_0\|_{2}^2 \right] + \eta KL\big(\mathcal{N}(\mu_y,\sigma_y)\|\mathcal{N}(0,1)\big)
\end{align*}
\vspace{-4 mm}
%


At test time, latent code vectors are sampled from a normal distribution $z \sim \mathcal{N}(\mu_x,\sigma_x)$ to generate new reconstructions. To ensure that these reconstructions do not deviate from physical measurement, data consistency (obtained by applying an affine projection based on the undersampling mask) is applied to all network outputs, which we find essential to obtaining high SNR \cite{mardani2018deep}. Fig. \ref{fig:architecture} depicts all of the model's components.

To deepen our analysis of robustness, we also examine the effects of a revised model architecture that is cascaded (i.e. the VAE and data consistency portions of the model repeat) for a certain number of "recurrent blocks," since prior work has shown that this can positively impact reconstruction quality \cite{mardani2018neural}. In the case of a model with two recurrent blocks, for example, the zero-filled image is passed into the first VAE, the output of the first VAE is passed as the input of a second VAE, and the output of the second VAE serves as the overall model reconstruction. Data consistency is applied to each VAE in question, and the VAEs have shared weights which ensures that new model parameters are not introduced. Note that we define a model with one recurrent block as the baseline model that has no repetition.




\section{Uncertainty Analysis}



The advantage of using a VAE model for image reconstruction is that it naturally offers the posterior of the image which can be used to draw samples and generate variance maps. While these variance maps are useful in letting radiologists understand which pixels differ the most between the model's reconstructions, they are not sufficient since the bias (difference between reconstructions and the fully-sampled ground truth image) is unknown. Thus, methods like SURE which can assess risk without explicitly using the ground-truth are a better alternative. The following subsections introduce these ideas in more detail.

\subsection{Monte Carlo sampling}

An effective way of analyzing the uncertainty of probabilistic models in computer vision problems is to statistically analyze output images for a given input \cite{kendall2017uncertainties}. Nonetheless, this approach has not been widely employed in inverse problems or medical image recovery. 

Utilizing the probabilistic nature of the VAE model, for a given zero-filled image $x_{zf}$ (i.e. the aliased input to the model), we can use our encoder function $f$ to find the mean $\mu$ and variance $\sigma^2$ of the latent code $z$, which we use to draw samples. We can then use our decoder function $g$ to produce reconstructions of latent code samples $z_i$ and then aggregate the results over $k$ samples to produce pixel-wise mean and variance maps for the reconstructions. Algorithm \ref{alg1} shows the full sequence of steps in more detail.

This Monte Carlo sampling approach allows one to evaluate variance as well as higher order statistics, which can be very useful in understanding the extent and impact of model uncertainty. However, despite the information the Monte Carlo approach can provide, some important statistics such as bias are dependent on knowledge of the ground truth, which motivates the use of metrics like SURE.

\begin{algorithm}[t]
	\caption{Monte Carlo uncertainty map } \small{
		
		\begin{algorithmic}
			\STATE \textbf{Input}  $y, \Omega, k, f, g$ 
			\STATE \textbf{Step 1.} Form the input $x_{zf} = \Phi^{\mathsf{H}}y$
			\STATE \textbf{Step 2.} Generate latent code statistics $(\mu, \sigma)=f(x_{zf})$ 
			\STATE \textbf{Step 3.} Initialize $i = 0$, $\hat{x}=[0]_{n \times n}$, and $map=[0]_{n \times n}$
			\STATE \textbf{Step 4.} While $i < k$  $ \{$ 
			\STATE \quad $i=i+1$
			\STATE \quad Draw latent code sample: $z_i \sim \mathcal{N}(\mu,\sigma^2)$
		    \STATE \quad Decode: $\hat{x}_i = g(z_i)$ 
		    \STATE \quad Collect reconstructions: $\{\hat{x}_i\}$
		    \STATE \quad $\}$
			\STATE \textbf{Step 5.} Compute pixel-wise mean and variance: $\hat{x}:=\frac{1}{k} \sum_{i=1}^k \hat{x}_i$, $map:=\frac{1}{k} \sum_{i=1}^k (\hat{x}_i-\hat{x})^2$ \\
			\vspace{0.5 mm}
			\RETURN ($\hat{x}, map$)
			
		\end{algorithmic}}
		\label{alg1}
	\end{algorithm}

\subsection{Denoising SURE}
A useful statistical technique for risk assessment when the ground truth is unknown is Stein's Unbiased Risk Estimator (SURE) \cite{tibshirani2018excess}. Despite being well-established, SURE has not yet been used for uncertainty analysis in imaging or DL problems. 

Given the ground truth image $x_0$, the zero-filled image can be written as
\begin{align}
    x_{zf} = x_0 + v
\end{align}
where $v$ is noise. Now considering reconstruction $\hat{x}$ with dimension $n$, one can expand test MSE as
\begin{align*}
\mathbb{E} \|\hat{x} - x_0\|^{2} & = \mathbb{E}\|x_0 - x_{zf} + x_{zf} - \hat{x} \|^{2} \\ & = -n\sigma^2 + \mathbb{E} \|x_{zf}-\hat{x} \|^2 + 2{Cov}(x_{zf},\hat{x})
\label{eq:n3}
\end{align*}
Since $x_0$ is not present in this equation, we see that SURE serves as a surrogate for MSE even when the ground truth is unknown. A key assumption behind SURE is that the noise process $v$ that relates the zero-filled image to the ground truth is normal, namely $v \sim \mathcal{N}(0,\sigma^2I)$. With this assumption, we can apply Stein's formula which approximates the covariance to obtain an estimate for the risk as
\begin{equation} 
SURE = -n\sigma^2 + \underbrace{\|\hat{x}-x_{zf} \|^2}_{\text{RSS}} + \sigma^2\underbrace{{tr}(\frac{\partial \hat{x}}{\partial x_{zf}})}_{\text{DOF}}
\label{eq:Sure_eq}
\end{equation} 
Note that SURE is \textit{unbiased}, namely
\begin{align}
    MSE = \mathbb{E}[SURE]
\end{align}
With the risk expressed in the above form, we can separate the estimate into three terms, with the second term corresponding to residual sum of squares (RSS) and the third one corresponding to degrees of freedom (DOF). This form importantly does not depend on $x_0$, and approximates the DOF with the trace of the end-to-end network Jacobian $J=\partial \hat{x}/\partial x_{zf}$. The Jacobian represents the network sensitivity to small input perturbations and is a measure of interest when analyzing robustness in computer vision tasks~\cite{jakubovitz2018improving}.

In order to estimate the noise variance, it is reasonable to assume that error in the output reconstruction is not large, and as a result $\sigma^2$ can be estimated (by setting the sum of the first two terms in the SURE expression to zero) as

\begin{equation} 
\sigma^2 = \|\hat{x}-x_{zf} \|^2 / n
\label{eq:Sigma_eq}
\end{equation} 

Our evaluations validate this assumption when the undersampling rate is not very large. With this assumption, we can rewrite our expression for SURE as follows

\begin{equation} 
SURE = \sigma^2{\rm tr}(\partial \hat{x} / \partial x_{zf})
\label{eq:Sure_eq_final}
\end{equation} 

\noindent\textbf{DOF approximation.}~ Due to the high dimension of MR images, computing the end-to-end network Jacobian for SURE can be computationally intensive. From a practical standpoint, making this computation efficient allows one to evaluate uncertainty in real time, in parallel with the reconstruction. To this end, we use an approximation for the Jacobian trace \cite{metzler2018unsupervised, ramani2008monte}. In particular, given $n$-dimensional noise vector $b$ drawn from $\mathcal{N}(0,1)$, our trained model $h$, the zero-filled model input $x_{zf}$, and a small value $\epsilon$ (which we define as the maximum pixel value in the zero-filled image divided by 1000), the approximation can be expressed as follows   

\begin{align} 
{\rm tr}\{J\} \approx b^T(h(x_{zf} + \epsilon b) - h(x_{zf}))\epsilon^{-1}
\label{eq:approx}
\end{align} 


\begin{figure*}
\begin{center}
\includegraphics[width = 18 cm] {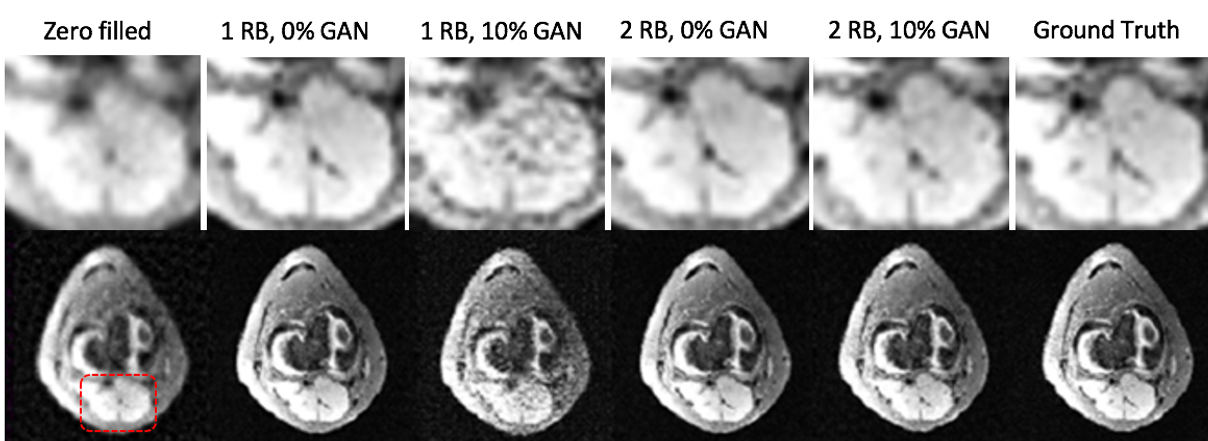}
\caption{\small The aliased input, reconstructions with one recurrent block (RB) and pure MSE loss, one RB and 10 percent GAN loss, two RBs and pure MSE loss, and two RBs and 10 percent GAN loss, and the ground truth for a representative slice.  }
\label{fig:recon_ims}
\end{center}
\end{figure*}

\subsection{Gaussian noise with density compensation}
As mentioned before, the key assumption behind SURE is that the noise model is Gaussian with zero mean. However, it is not safe to assume that the undersampling noise in MRI reconstruction inherits this property. For this reason, we introduce density compensation on the input image as a way of enforcing zero-mean residuals. This approach has the added benefit of making artifacts independent of the underlying image and we find that it significantly increases residual normality (see Fig. \ref{fig:qq_density} in Section 6).

More specifically, given a 2D sampling mask $\Omega$, we can treat each element of the mask as a Bernouli random variable with a certain probability $D_{i,j}$, where $\mathbb{E}[\Omega]=D$ (this is dependent on the sampling approach).

With this formulation, we can define a density compensated zero-filled image as follows: $\tilde{x}_{zf}= F^{-1}{D}^{-1} \Omega F x_0$. We can rewrite this expression using $x_0$ as

\begin{align}
   \tilde{x}_{zf} = x_0 + \underbrace{(F^{-1}{D}^{-1} \Omega F - I) x_0}_{:=v}   \label{eq:sure_density_comp}
\end{align}

First, we observe that $v$ has zero mean since $\mathbb{E}[{D}^{-1} \Omega ]=I$. In addition, the noise variance obeys
\begin{align}
    \sigma^2 = {\rm tr} \big( x_0^{\mathsf{H}} (F^{-1}{D}^{-1} F - I) x_0 \big)
    \label{eq:sigma}
\end{align}

Of course, in practice we do not have access to the ground truth image $x_0$, and instead rely on the approximation in \ref{eq:Sigma_eq} for the noise variance. Given these main properties, the density compensation method that this work introduces represents an important step that can allow denoising SURE to be used effectively in medical imaging and other inverse problems. Algorithm \ref{alg2} summarizes the steps for using density-compensated SURE in practice.

\begin{algorithm}[t]
	\caption{SURE risk estimate (with density compensation)} \small{
		
		\begin{algorithmic}
		
			\STATE \textbf{Input}  $y, \Omega, D, n, h, b, \epsilon$ 
			\STATE \textbf{Step 1.} Form density-compensated input  $\tilde{x}_{zf}= F^{-1}{D}^{-1} \Omega F \Phi^{\mathsf{H}} y$

			\STATE \textbf{Step 2.} Reconstruct $\hat{x} = h(\tilde{x}_{zf})$

            \STATE \textbf{Step 3.} Compute the noise variance
            			\begin{equation*} 
            \sigma^2 = \|\hat{x}-\tilde{x}_{zf} \|^2 / n
            \end{equation*} 
            
			\STATE \textbf{Step 4.} Approximate the DOF
			\begin{equation*} 
            {\rm tr}\{J\} \approx b^T(h(\tilde{x}_{zf} + \epsilon b) - h(\tilde{x}_{zf}))\epsilon^{-1}
            \end{equation*}

			\STATE \textbf{Step 5.} Obtain SURE risk for $\hat{x}$
			\begin{equation*} 
SURE = \sigma^2 {\rm tr}\{J\}
\end{equation*} 
\vspace{-4 mm}
			\RETURN{$SURE$}
		\end{algorithmic}}
		\label{alg2}
	\end{algorithm}

\vspace{2mm}

\noindent\textbf{Remark 1 [Compressed Sensing SURE]}.~The primary assumption behind the SURE derivations is that the noise obeys an i.i.d. Gaussian distribution. Even though the density compensation in \eqref{eq:sure_density_comp} enforces this property, one can also leverage the generalized SURE formulation that extends SURE derivations to colored noise distributions from the exponential family, though this is harder to compute in practice~\cite{deledalle2012unbiased, eldar2009generalized}. 

\begin{figure*}[t]
\begin{center}
\includegraphics[width = 11.5 cm] {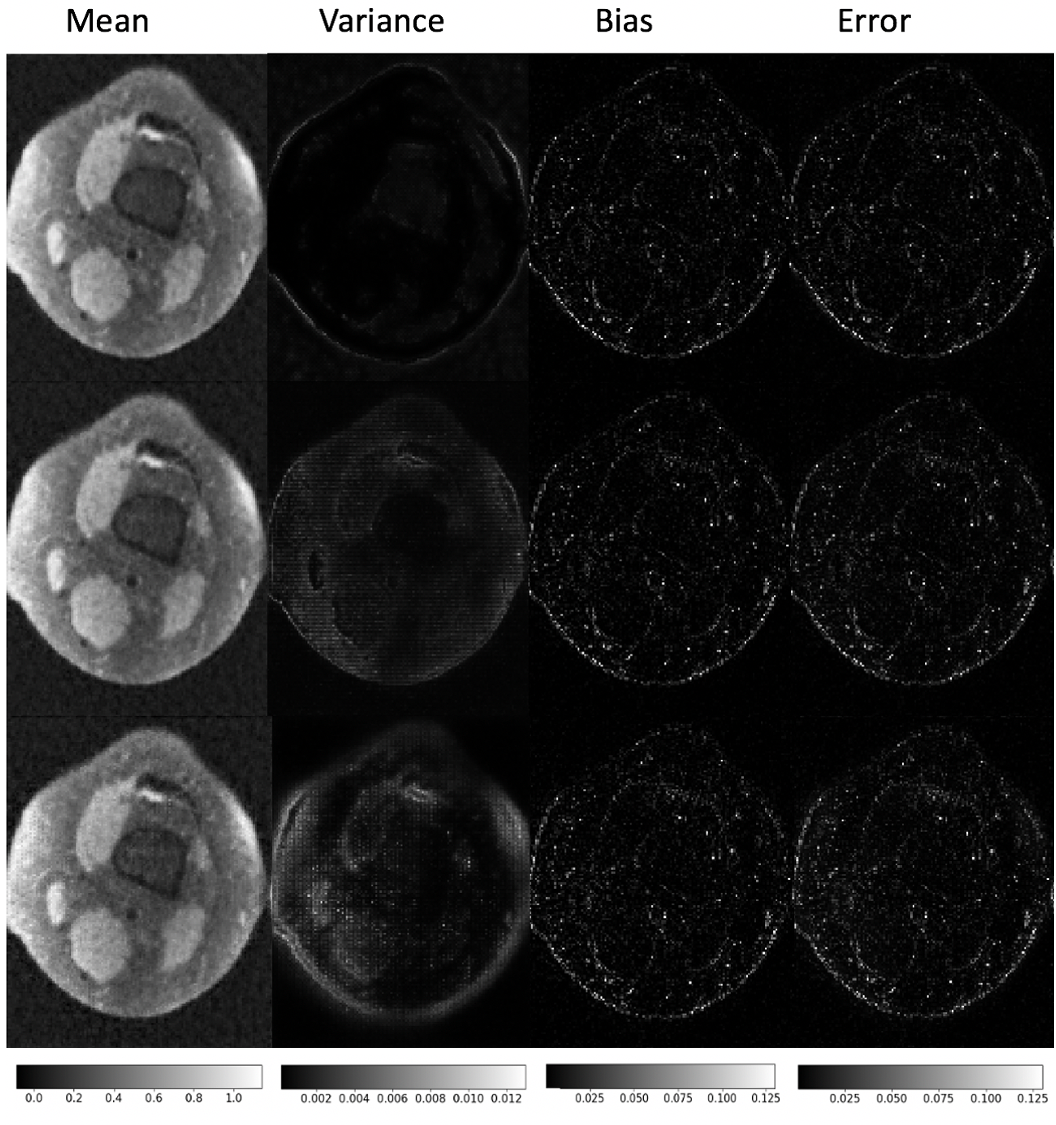}
\caption{\small Mean reconstruction, pixel-wise variance, bias, and error for a given reference slice across all realizations (5 fold undersampling and one recurrent block). Row 1: 0\% GAN loss ($\lambda = 0$) . Row 2: 5\% GAN loss ($\lambda = 0.05$). Row 3: 10\% GAN loss ($\lambda = 0.10$). }
\label{fig:uncertainty_gan}
  \end{center}
\end{figure*}

\begin{figure*}[t]
\begin{center}
\includegraphics[width = 11.5 cm] {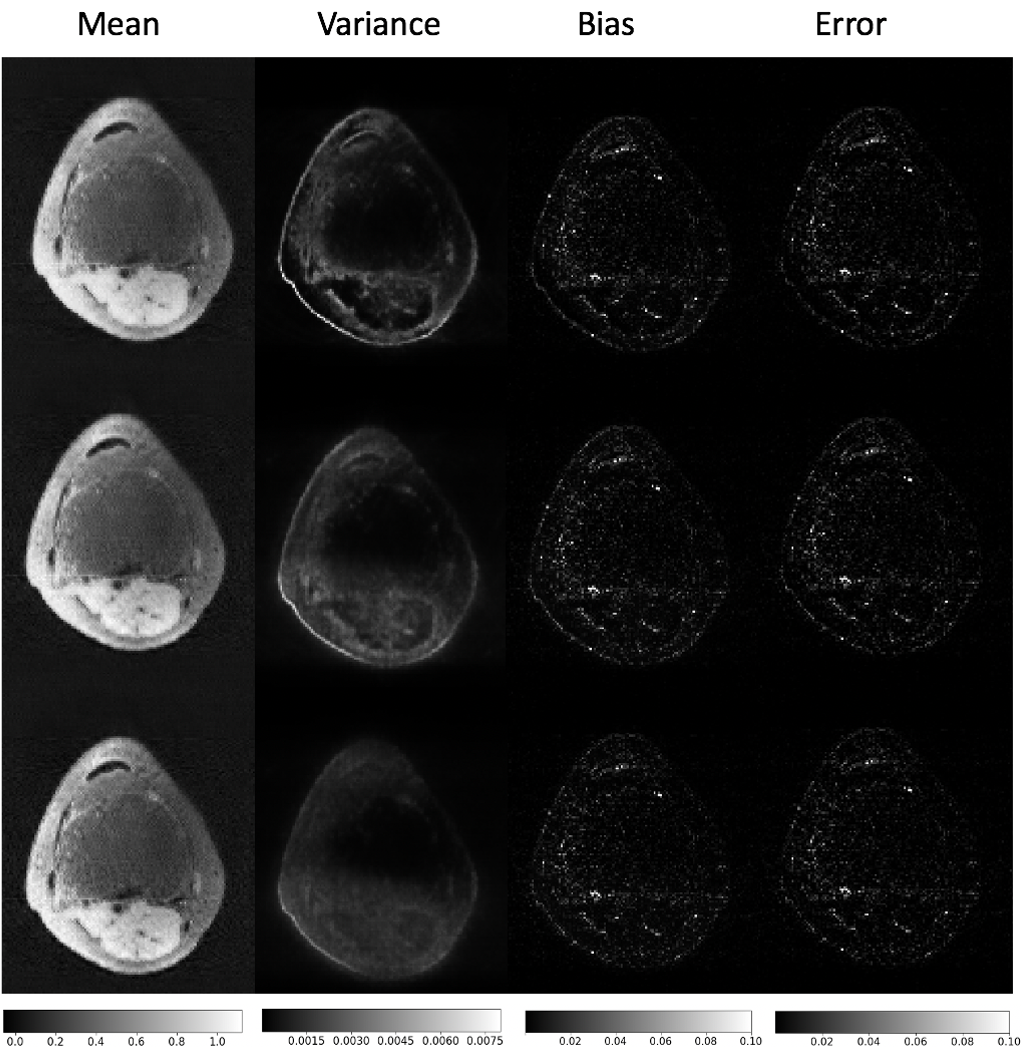}
\caption{\small Mean reconstruction, pixel-wise variance, bias, and error for a given reference slice across all realizations (5 fold undersampling and no adversarial loss). Row 1: one RB. Row 2: two RBs. Row 3: four RBs. }
\label{fig:uncertainty_iter}
  \end{center}
\end{figure*}

\section{Empirical Evaluations}
In this section, we assess our model and methods on a dataset of Knee MR images. We first show reconstructions produced with the VAE model, before demonstrating representative results with the Monte Carlo and SURE methods for quantifying uncertainty.

\noindent\textbf{Dataset}.~The Knee dataset used for all experiments was obtained from $19$ patients with a 3T GE MR750 scanner \cite{mridata.org}. Each volume consists of 320 2D slices of dimension $320\times256$ that were divided into training, validation, and test examples with a 70/15/15 split stratified by patient. A 5-fold variable density undersampling mask with radial view ordering (designed to preserve low-frequency structural elements) was used to produce aliased input images $x_{zf}$ for the model to reconstruct \cite{cheng2013variable}.

\subsection{Adversarial Loss}

The effective modeling of high-frequency components is essential for capturing details in medical images. Thus, we train our model with adversarial loss in a GAN setup, which can better capture high-frequency details \cite{mardani2018deep, larsen2015autoencoding}. In particular, we use a multi-layer CNN as the discriminator $\mathcal{D}$ along with the VAE generator $\mathcal{G}$. The discriminator learns to distinguish between reconstructions and fully-sampled images, and sends feedback to the generator, which in turn adjusts the VAE's model weights to produce more realistic reconstructions.

The training process for the VAE is identical to the case with no adversarial loss, except an extra loss term is needed to capture the discriminator feedback (with weight $\lambda$ referred to as GAN loss) 

\vspace{-1mm}
\begin{equation*}
\begin{split}
\min_{G}     \mathbb{E}_{x,y} \left[ \|\hat{x} - x_0\|_{2}^2 \right] + \eta KL\big(\mathcal{N}(\mu_y,\sigma_y)\|\mathcal{N}(0,1)\big)  \\ + \lambda \mathbb{E}_{y}\left[(1 - \mathcal{D}(\hat{x}))^{2} \right]
\label{eq:gan_d}
\end{split}
\end{equation*}

The discriminator weights are updated during training as
\vspace{0 mm}
\begin{equation*} 
 \min_{D} \mathbb{E}_{x} \left[(1-\mathcal{D}(x))^{2} \right] + 
 \mathbb{E}_{y} \left[(\mathcal{D}(\hat{x}))^{2} \right]
\label{eq:gan_gen}
\end{equation*} 

The training process acts as a game, with the generator continuously improving its reconstructions and the discriminator distinguishing them from ground truth. As the GAN loss $\lambda$ increases, the modeling of realistic image components is enhanced but uncertainty rises.

\subsection{Network architecture}

As shown in Fig. \ref{fig:architecture}, the VAE encoder is composed of $4$ convolutional layers ($128$, $256$, $512$, and $1024$ feature maps, respectively) with kernel size $5 \times 5$ and stride $2$, with each followed by ReLU activations and batch normalization \cite{van2017neural}. Latent space mean, $\mu$, and standard deviation, $\sigma$, are each represented by fully connected layers with $1024$ neurons. The VAE decoder has $5$ layers and utilizes transpose convolution operations ($1024$, $512$, $256$, $128$, and $2$ feature maps, respectively) with kernel size $5 \times 5$ and stride $2$ for upsampling \cite{jegou2017one}. Skip connections are utilized to improve gradient flow through the network \cite{dieng2018avoiding}. 

The discriminator function of the GAN (when adversarial loss is used) is an 8-layer CNN. The first seven layers are convolutional ($8$, $16$, $32$, $64$, $64$, $64$, and $1$ feature maps, respectively) with batch normalization and ReLU activations in all but the last layer. The first five layers use kernel size $3 \times 3$, while the following two use kernel size $1 \times 1$. The first four layers use a stride of $2$, while the next three use a stride of $1$. The eighth and final layer averages the output of the seventh layer to form the discriminator output.

The use of multiple recurrent blocks (RB) whereby the model repeats (the output feeds into the input of another VAE with the same model parameters) is also explored \cite{mardani2018neural}. Using multiple RBs does not affect the discriminator network architecture.

Training was completed over the course of 30K iterations, with loss converging over roughly 20K iterations. We utilize the Adam optimizer with a mini-batch size of 4, an initial learning rate of $5 \times 10^{-5}$ that was halved every 5K iterations, and a momentum parameter of 0.9. Models and experiments were developed using TensorFlow on an NVIDIA Titan X Pascal GPU with 12 GB RAM. A version of our TensorFlow source code is publicly available via GitHub \cite{github}.


\subsection{Individual reconstructions}
Fig.~\ref{fig:recon_ims} shows sample model reconstructions (using the mean of the output distribution) for representative test slices with different hyperparameters. As the number of RBs increases from one to two (columns 2 and 3 versus 4 and 5), the resulting outputs improve in quality (corresponding to a roughly 1 dB gain in SNR). Additionally, progressively increasing values of GAN loss (columns 2 and 4 versus 3 and 5) introduce high-frequency artifacts to the image, while leading to sharper outputs. The highlighted ROI elucidates these effects, where the degradation in visual quality associated with poor recovery can be detrimental to radiologist diagnoses. As expected, the presence of substantial adversarial loss decreases average reconstruction SNR (and increases MSE) as Table \ref{tab2} shows. With limited GAN loss and additional RBs, though, the SNR is close to 20 dB, indicating that the probabilistic model results in effective image recovery, in addition to facilitating uncertainty analysis.

\subsection{Variance, bias, and error maps}

Using the Monte Carlo method described earlier in Algorithm \ref{alg1}, 1K outputs corresponding to different reference slices were generated after feeding a test image into the trained model and sampling from the resulting posterior distribution. Note that for this process, only the VAE (generator portion) of the model is relevant to producing outputs, even with adversarial loss used for training.

We show the mean of the 1K reconstructed outputs for a representative slice, the pixel-wise variance, bias (difference between mean reconstruction and ground truth), and error in Figs. \ref{fig:uncertainty_gan} and \ref{fig:uncertainty_iter}, utilizing the common relation $error = bias^{2} + variance$ \cite{tibshirani1996bias}. The concept of bias and variance is important in the analysis of uncertainty because both the difference from the ground truth and the inherent variability across realizations provide information on the portions of a given image most susceptible to the introduction of realistic artifacts. Note that to compute the bias and error, we need the ground-truth which is provided here solely for validation purposes.

The results indicate that variance, bias, and error increase as the GAN loss weight $\lambda$ increases (Fig. \ref{fig:uncertainty_gan}) and as the number of RBs decreases (Fig. \ref{fig:uncertainty_iter}). Furthermore in all cases with GAN loss, the variance extends to structural components of the image, which poses the most danger in terms of diagnosis. Nevertheless, with a reasonably conservative choice of GAN weight, the risk is substantially lower. More RBs can lower variance as well as error, and can be a useful tweak to improve robustness. Although only a few representative examples are shown here, similar trends were observed with all examined reference slices.

\begin{figure*}
\begin{center}
\includegraphics[width = 15.8 cm] {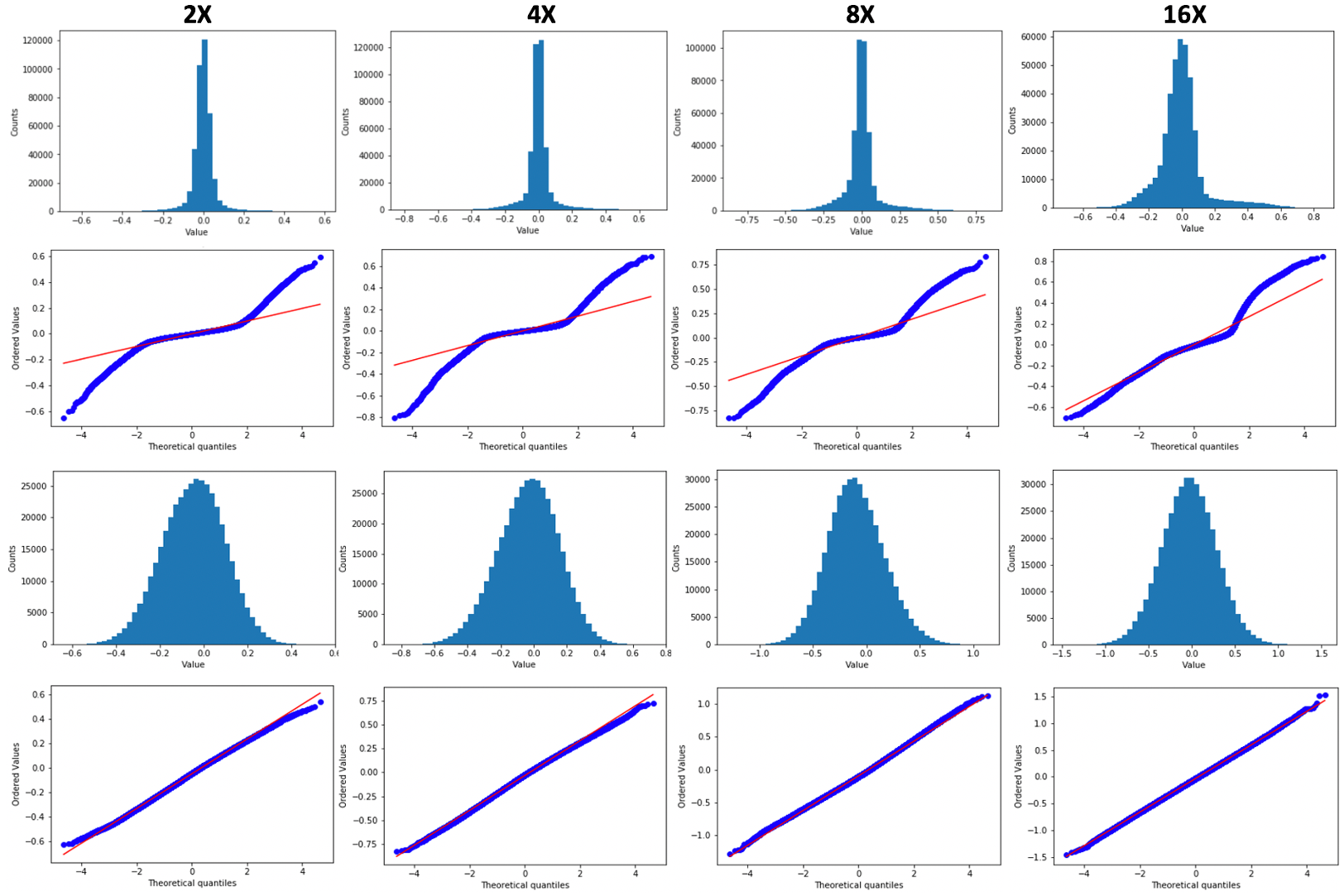}
\caption{\small Histograms and quantile-quantile plots of the residuals between zero-filled and ground truth images without (top two rows) and with density compensation with 2, 4, 8, and 16 fold undersampling, respectively. }
\label{fig:qq_density}
  \end{center}
\end{figure*}

\begin{figure*}
\begin{center}
\includegraphics[width = 18 cm] {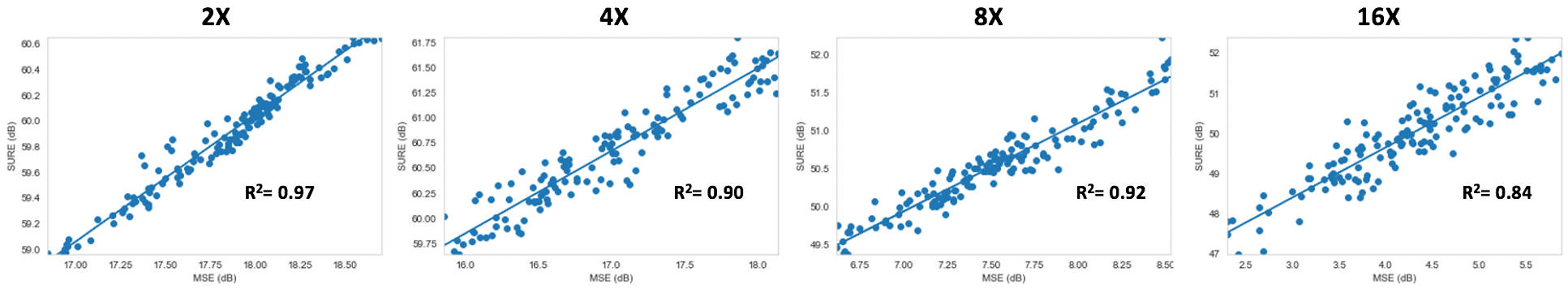}
\caption{\small SURE vs. MSE for one recurrent block using no adversarial loss, with 2-fold, 4-fold, 8-fold, and 16-fold undersampling. }
\label{fig:corr}
  \end{center}
\end{figure*}

\subsection{Noise distribution with density compensation}
As described earlier, denoising SURE builds on the Gaussian noise assumption for the residuals $v = x_{zf} - x_0$. To validate this assumption, we produce histograms and Q-Q plots of the residuals at various undersampling rates, by considering the differences for individual pixels across test images. From the top two rows of Fig. \ref{fig:qq_density} one can observe the quantiles closer to the center of the distribution are aligned with those of the normal distribution for all undersampling rates. The noise distribution is not perfectly normal in any of the cases, however, which can limit the effectiveness and accuracy of SURE.  


To overcome the lack of normality in the residuals, we apply our density compensation method (section V.C). For a given undersampling rate, 100 variable-density random sampling masks are generated and then averaged to obtain the sampling density $D$. The element-wise inverse of this density can then be multiplied with any given mask to produce a density-adjusted mask $D^{-1}\Omega$. This adjusted mask can be used to generate new zero-filled images as input to the network.

As Fig. \ref{fig:qq_density} (bottom two rows) shows, the distribution of residuals (for various undersampling rates) better matches the normal distribution, and the mean of the distribution lies very close to zero. This observation indicates density compensation is a valuable preprocessing step that enables the use of denoising SURE for quantifying the uncertainty in MRI reconstruction.

\begin{figure*}[t]
\begin{center}
\includegraphics[width = 12.5 cm] {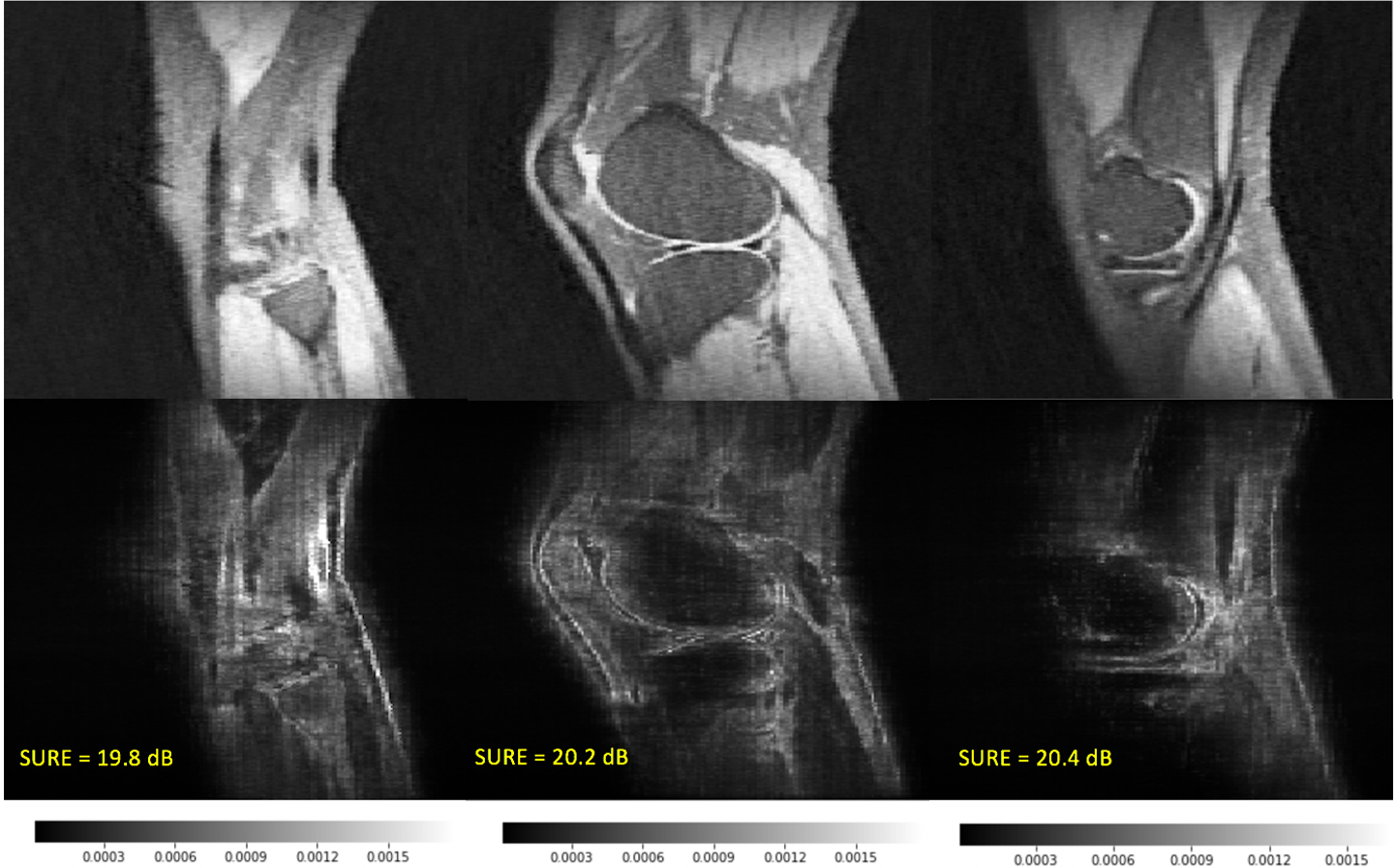}
\caption{\small Sagittal reconstructions (top row) along with corresponding variance maps (5 fold undersampling and one recurrent block with no adversarial loss) and average pixel risk (SURE) values.  }
\label{fig:sagittal}
  \end{center}
\end{figure*}

\begin{table}[t]
\caption{  Averaged results for different undersampling rates with one recurrent block and no adversarial loss}
\centering
\begin{adjustbox}{width= 7 cm}
\begin{tabular}{lllll}
\hline
\multicolumn{1}{|l|}{Metrics} & \multicolumn{1}{l|}{2-fold} & \multicolumn{1}{l|}{4-fold} & \multicolumn{1}{l|}{8-fold} & \multicolumn{1}{l|}{16-fold} \\ \hline
\multicolumn{1}{|l|}{SURE-MSE $R^2$} & \multicolumn{1}{l|}{0.97} & \multicolumn{1}{l|}{0.90} & \multicolumn{1}{l|}{0.92} & \multicolumn{1}{l|}{0.84} \\ \hline 
\multicolumn{1}{|l|}{MSE} & \multicolumn{1}{l|}{0.017} & \multicolumn{1}{l|}{0.021} & \multicolumn{1}{l|}{0.18} & \multicolumn{1}{l|}{0.39} \\ \hline
\multicolumn{1}{|l|}{RSS} & \multicolumn{1}{l|}{0.061} & \multicolumn{1}{l|}{0.065} & \multicolumn{1}{l|}{0.11} & \multicolumn{1}{l|}{0.13} \\ \hline
\multicolumn{1}{|l|}{DOF} & \multicolumn{1}{l|}{0.12} & \multicolumn{1}{l|}{0.16} & \multicolumn{1}{l|}{0.23} & \multicolumn{1}{l|}{0.29} \\ \hline
\multicolumn{1}{|l|}{SNR (dB)} & \multicolumn{1}{l|}{20.7} & \multicolumn{1}{l|}{19.1} & \multicolumn{1}{l|}{16.8} & \multicolumn{1}{l|}{15.5} \\ \hline
\multicolumn{1}{|l|}{SURE (dB)} & \multicolumn{1}{l|}{21.3} & \multicolumn{1}{l|}{19.8} & \multicolumn{1}{l|}{15.9} & \multicolumn{1}{l|}{14.2} \\ \hline

\end{tabular}
\end{adjustbox}
\label{tab1}
\end{table}

\begin{table*}[t]
\caption{ Averaged results for different quantities of recurrent blocks and adversarial loss with 5-fold undersampling}
\centering
\begin{adjustbox}{width= 12 cm}
\begin{tabular}{lllll}
\hline
\multicolumn{1}{|l|}{Metrics} & \multicolumn{1}{l|}{1 RB, 0\% GAN} & \multicolumn{1}{l|}{1 RB, 10\% GAN} & \multicolumn{1}{l|}{2 RB, 0\% GAN} & \multicolumn{1}{l|}{2 RB, 10\% GAN} \\ \hline
\multicolumn{1}{|l|}{SURE-MSE $R^2$} & \multicolumn{1}{l|}{0.95} & \multicolumn{1}{l|}{0.88} & \multicolumn{1}{l|}{0.95} & \multicolumn{1}{l|}{0.91} \\ \hline
\multicolumn{1}{|l|}{MSE} & \multicolumn{1}{l|}{0.033} & \multicolumn{1}{l|}{0.082} & \multicolumn{1}{l|}{0.027} & \multicolumn{1}{l|}{0.057} \\ \hline
\multicolumn{1}{|l|}{RSS} & \multicolumn{1}{l|}{0.092} & \multicolumn{1}{l|}{0.12} & \multicolumn{1}{l|}{0.076} & \multicolumn{1}{l|}{0.11} \\ \hline
\multicolumn{1}{|l|}{DOF} & \multicolumn{1}{l|}{0.17} & \multicolumn{1}{l|}{0.20} & \multicolumn{1}{l|}{0.16} & \multicolumn{1}{l|}{0.17} \\ \hline
\multicolumn{1}{|l|}{SNR (dB)} & \multicolumn{1}{l|}{18.8} & \multicolumn{1}{l|}{17.1} & \multicolumn{1}{l|}{19.9} & \multicolumn{1}{l|}{17.9} \\ \hline
\multicolumn{1}{|l|}{SURE (dB)} & \multicolumn{1}{l|}{18.0} & \multicolumn{1}{l|}{16.2} & \multicolumn{1}{l|}{19.2} & \multicolumn{1}{l|}{17.3} \\ \hline

\end{tabular}
\end{adjustbox}
\label{tab2}
\end{table*}

\vspace{-0.5 mm}
\subsection{SURE results}

To evaluate the effectiveness of the density-compensated SURE approach, we produce correlations of SURE versus MSE (which depends on the ground truth and is a standard metric for assessing model error) using the results from our test images. Fig. \ref{fig:corr} shows the strong linearity of the correlations under all conditions. The linear relationship is strongest for lower undersampling rates ($R^2 = 0.97$ for 2-fold while $R^2 = 0.84$ for 16-fold). Nonetheless, the results show that even with relatively high undersampling, SURE can be used to effectively estimate risk in medical image reconstructions.

In addition, the average reconstruction SURE, RSS, and DOF values for different hyperparameters are shown in Table \ref{tab1} and Table \ref{tab2}. Increased GAN loss results in decreased SURE values (note the units of dB) and increased RSS and DOF. Meanwhile, more RBs result in higher SURE values and lower RSS and DOF, demonstrating a simple way of improving reconstruction quality while reducing uncertainty. Note that these results align closely with the Monte Carlo analysis from before, thereby reinforcing the effectiveness of the SURE approach in quantifying risk. 

Furthermore, the uncertainty analysis of the sample reconstructions in Fig. \ref{fig:sagittal} indicates that there is a relationship between SURE values and variance magnitude. As overall variance decreases from left to right, the SURE values (again, in dB) increase correspondingly. In practice, these two methods need to be used in conjunction, with the variance maps illustrating the localization of uncertainty and SURE providing a more statistically sound global metric.

\vspace{-0.5 mm}
\section{Conclusions}
%
This paper introduces methods to analyze uncertainty in deep-learning based compressive MR image recovery. To thoroughly explore realistic and data-consistent images, we develop a probabilistic VAE model with low-error. A Monte Carlo approach is used to quantify the pixel variance and obtain uncertainty maps in parallel with reconstruction. Moreover, to fully assess the reconstruction risk (which requires the bias and thus fully sampled data), we leverage Stein's Unbiased Risk Estimator on density-compensated zero-filled images where the noise obeys zero-mean Gaussian. We validate the utility of these tools with evaluations on a dataset of Knee MR images. 

In particular, our observations demonstrate that increased adversarial loss leads to larger uncertainty, indicating a trade-off when using adversarial loss to better retrieve high-frequencies. On the other hand, using multiple recurrent blocks (cascaded VAE and data consistency layers), decreases uncertainty, which suggests an effective way of promoting robustness.

There are still important future directions to explore. While we focus on model uncertainty in this work, there are other sources of uncertainty (pertaining to data and knowledge) that should be taken into account. As such, next steps involve extensive evaluations with different MRI datasets and acquisition strategies to assess the effects of data uncertainty on model reconstructions. Additionally, it would be useful to perform uncertainty analysis for pathological cases. Quantifying the likelihood of a diagnosis being altered by hallucinated artifacts, and finding regularization schemes to limit this, will improve the effectiveness of DL methods for MRI reconstruction.


Finally, it would be valuable to extend  SURE to obtain local, pixel-wise risk in addition to a global measure pertaining to the overall image. The pixel-wise SURE could then be used to generate uncertainty maps that incorporate valuable spatial information, similar to Monte-Carlo variance maps.


\newpage






%
\bibliography{main}
\bibliographystyle{IEEEtran}
\end{document}